\documentclass{llncs}

\usepackage{comment}
\usepackage{graphicx}
\usepackage{multirow}
\usepackage{subcaption}
\usepackage{amsmath}
\usepackage{todonotes}

\begin{document}

\title{Benchmark of visual and 3D lidar SLAM systems in simulation environment for vineyards}
%\todo{either capitlise everything or make visual and lidar small case.}
%\title{Evaluation of Visual and 3D Lidar SLAM methods in virtual vineyard}

%\author{Third Draft}

\author{Ibrahim Hroob \and Riccardo Polvara \and Sergi Molina \and Grzegorz Cielniak \and Marc Hanheide }

\institute{
Lincoln Center for Autonomous Systems, University of Lincoln, UK\\
\email{ \{ihroob,rpolvara,smolinamellado,gcielniak,mhanheide\}@lincoln.ac.uk} }

\maketitle

\begin{abstract}

In this work, we present a comparative analysis of the trajectories estimated from various Simultaneous Localization and Mapping (SLAM) systems in a simulation environment for vineyards. Vineyard environment is challenging for SLAM methods, due to visual appearance changes over time, uneven terrain, and repeated visual patterns. For this reason, we created a simulation environment specifically for vineyards to help studying SLAM systems in such a challenging environment. We evaluated the following SLAM systems: LIO-SAM, StaticMapping, ORB-SLAM2, and RTAB-MAP in four different scenarios. The mobile robot used in this study equipped with 2D and 3D lidars, IMU, and RGB-D camera (Kinect v2). The results show good and encouraging performance of RTAB-MAP in such an environment. 


\end{abstract}
%\todo{You can add a note on general applicability of the presented methods in outdoor scenarios.} 

\section{Introduction} \label{introduction}

%notes: talk about gbs and how it is used for localization, then about no gbs what are the issues, current solution , comparison, what I will add to the party 

%\todo{not sure how facing the shortages is connected to better management of the farm, remove? the second part covers labour aspect sufficiently}
%\todo{better to use RTK-GNSS (more general) than GPS (US-based implementation of GNSS)}
%\todo{so how is this an issue? state that this requires a good coverage of base stations to guarantee the accuracy}
Precision agriculture relies on collecting data from multiple sensors to help improving farm management and crop yield. Better management of the farm requires continuous monitoring of the plant health and soil condition, discovering diseases at an early stage and reducing chemical treatment. To achieve these goals, one solution would be to use a mobile robot to autonomously inspect the plants and the crops. In that context, the mobile robot must have an accurate representation of the farm to accurately localize itself and navigate to the goals. For this reason, many solutions have been proposed to overcome the localization problem in an outdoor environment. The most common solution is to use Real-Time Kinematic Global Positioning Systems (RTK-GNSS)~\cite{Nrremark2008}, however, this solution is quite expensive and requires good coverage of base stations to guarantee accuracy. Other solutions rely on consumer-grade GNSS with fusing the output with different onboard sensors to enhance localization accuracy \cite{Imperoli2018}. 

%\todo{GNSS}
%\todo{it would be good to add a reference here, perhaps from those that you already have.}

GNSS is not always available and the signal may not be reliable due to environmental conditions; loss of signal for the autonomous robot may lead to catastrophic failures. Therefore, alternative solutions have been proposed based on the Simultaneous Localization and Mapping (SLAM) concept \cite{Aguiar2020}, where robot pose is estimated using sensory input, while at the same time building map of the environment. Various SLAM systems have been developed in that regard \cite{labbe_2019}. The selection of a suitable SLAM system depends on multiple factors, such as type (indoor or outdoor) and scale of the environment. Another factor is the sensors' cost, for example, some systems rely on data from relatively expensive 3D lidars whilst others use data from cheap consumer-grade monocular cameras. However, most of the developed solutions were targeting either the indoor environments or outdoor urban environments \cite{Aguiar2020}. Nevertheless, SLAM in agricultural applications is still a growing field due to challenges related to harsh environmental conditions, seasonal changes in appearance, and repeated visual features in large open fields.

The main contributions of this paper are (i) Releasing to the public an open-source realistic vineyard simulator, offering uneven terrain and five different stages of plant growth\footnote{github.com/LCAS/bacchus\_lcas}. (ii) Comparing and benchmarking 4 SLAM systems in an environment with repeating structure and appearance. The algorithms chosen for this study represent the sate-of-the-art visual and 3D-lidar systems including LIO-SAM \cite{liosam2020shan}, StaticMapping \cite{edward_liu_2021}, ORB-SLAM2 \cite{MurArtal2017}, and RTAB-MAP \cite{labbe_2019}.

%Those system are consider the state-of-the-are in V-SLAM and 3D lidar SLAM at the time of writing this paper. 

%I had to comment this paragraph due to limited space and it does not add any value
%The work in this paper is divided as follows: In Section \ref{related_work} we presented the related work in comparing SLAM systems. Section \ref{slam_algorithms} is a brief summary of the used SLAM systems in this paper. The evaluation and result discussions are in Section \ref{evaluation}. Finally, we concluded this work in Section \ref{conclusion}.
% I may need to stress what is the ground truth and how the researchers used it may be!? 
% visual 2d 3d actual data set 
% hmm I think it will be good idea to make this section follow a story or time line of the developed systems? 
% The lack of work done in comparing 3D lidar SLAM systems 
% little studied in the literature

\section{Related work} \label{related_work}
There has not been much work done in comparing various SLAM systems specifically for the vineyard environment, nevertheless, there have been many research papers dealing with analyzing and comparing SLAM methods for indoor static environments. In this section, we review the most relevant papers in this area. 

The authors in \cite{Filipenko2018} evaluated the trajectory generated from different ROS-based SLAM algorithms in a typical office indoor environment. The mobile robot was equipped with a 2D laser scanner, a monocular and stereo camera. The evaluation was on a specifically acquired data-set. The authors used the estimated trajectory from the best performing 2D lidar SLAM as the ground truth for visual SLAM systems. The results were good and encouraging for RTAB stereo with Root Mean Square Error (RMSE) of 0.163 m, and for ORB-SLAM monocular with RMSE of 0.166 m. %The researchers in \cite{Yagfarov2018} use a precise laser tracker for accurate ground truth to compare the three most common 2D SLAM algorithms, gmapping, hector\_slam, and google cartographer. The mobile robot was equipped with a 2D laser scanner whilst the evaluation considered the accuracy of map construction and trajectory generated. The results showed that in this particular scenario, Google Cartographer is the most accurate algorithm compared to others.

Part of the work in \cite{labbe_2019} is an evaluation of the trajectory performance between different sensor configuration of RTAB-MAP (stereo and lidar), LSD-SLAM (stereo), ORB-SLAM2 (stereo) and SOFT-SLAM (stereo) in the outdoor KITTI dataset \cite{Geiger2012}. At this dataset, the authors stated both lidar and stereo configuration have followed well the ground truth. However, in some sequences with not a complex structure, stereo setup systems outperform the lidar configuration of RTAB-MAP. Stereo RTAB with ORB feature detection has performed well in 9 out of 11 sequences, but that configuration was computationally expensive and was not able to meet real-time constraint on their hardware setup. 

An Extended Information Filter (EIF) was used for mapping an agricultural environment and localizing a mobile robot \cite{AuatCheein2011}. The system makes use of a 2D laser scanner and monocular camera to detect olive tree stems. The tests were done in a real agricultural environment. However, the authors stated that they find some errors in identifying features and detecting loop closure. In \cite{Comelli2019} the authors evaluated visual SLAM methods, such as ORB-SLAM2 and S-PTAM\footnote{github.com/lrse/sptam}, against visual-inertial SLAM system as S-MSCKF\footnote{github.com/KumarRobotics/msckf\_vio} on the Rosario dataset \cite{Pire2019}. The result showed poor accuracy and robustness compared to an indoor or urban environment, where those algorithms are designed for. Another study \cite{Capua2018} comparing three visual SLAM algorithms in an orchard of fruit trees. The results showed that ORB-SLAM2 is the most accurate system with its loop closure ability. 

Based on the above literature, there is a lack of comparison between SLAM methods in vineyards environment. Therefore, more research needs to be done in that field.

%\todo[inline]{You should add the benchmarks for autonomous cars as this is more relevant than indoor scenarios: and specifically the KITTI benchmark.}
\section{SLAM Algorithms} \label{slam_algorithms}
% I need to add ros thing, which algorithm is integrated in ros and which one has a ros wrapper around it 
% also I need to add how I configure each one with the environment 
The SLAM systems used in this work can be classified into two main categories: visual SLAM and 3D lidar SLAM. This section briefly describes the four tested algorithms, RTAB-MAP and ORB-SLAM2 under visual SLAM, LIO-SAM, and StaticMapping under 3D lidar SLAM. Those systems have been chosen since they are the state-of-the-art, and the most popular algorithms by the time writing this paper. However, StaticMapping is not very popular yet. A brief list of sensors employed by each system is shown in Table \ref{table:slam_table} including the front-end back-end algorithms used by each method. The sensors that are used by each system in our setup are marked as bold text in the table. 

%\todo{Can you state why you have chosen those four? These are state of the art but also most mature, perhaps most popular (e.g. check number of forks on github), but also provide the full system including back and front ends. Also Table 1 includes front and back ends - add this in text if this is right.}

%note I was not able to include the front end system, because it is very complicated for RTAB and ORB. For RTAB they are using multiple algorithms in the front end such as GFTT/BRIEF for feature extraction, NNDR for feature matching, RANSAC for motion estimation a local bundle adjustment for local optimization ... and that is only for the visual odom node, and there is another intresting algorithms for the lidar node ... 
\begin{table}[]
\caption{The SLAM systems used for benchmarking together with supported sensors, front end and back end algorithms.}
\label{table:slam_table}
\centering
\begin{tabular}{|c|c|c|c|}
\hline
\textbf{System} & \textbf{Sensors support} & \textbf{Front-end} & \textbf{Back-end} \\ \hline
\textbf{LIO-SAM} & \begin{tabular}[c]{@{}c@{}}\textbf{3D lidar} + \textbf{IMU},\\ GNSS (Optional)\end{tabular} & ICP & GTSAM  \\ \hline
\textbf{\begin{tabular}[c]{@{}c@{}}Static-\\ Mapping\end{tabular}} & \begin{tabular}[c]{@{}c@{}}\textbf{3D lidar}, IMU (Optional),\\ GNSS (Optional)\end{tabular} & ICP & GTSAM \\ \hline
\textbf{\begin{tabular}[c]{@{}c@{}}ORB-\\ SLAM2\end{tabular}} & \begin{tabular}[c]{@{}c@{}}Monocular,  \textbf{RGB-D} or\\ Stereo cameras\end{tabular} & \begin{tabular}[c]{@{}c@{}}ORB features extraction.\\ PnP RANSAC for motion\\ estimation.\end{tabular} & g2o \\ 
\hline \textbf{\begin{tabular}[c]{@{}c@{}}RTAB-\\ MAP\end{tabular}} & \begin{tabular}[c]{@{}c@{}}\textbf{RGB-D} or Stereo camera \\ (Mainsensors). \\ 2D or 3D lidar (Optional\\ to enhance map build from\\ main input sensors)\end{tabular} & \begin{tabular}[c]{@{}c@{}}Visual odom: GFTT/BRIEF\\ for feature detectoin, \\ NNDR for feature matching,\\ PnP RANSAC for motion\\ estimation.\\ lidar odom: ICP\end{tabular} & \begin{tabular}[c]{@{}c@{}}GTSAM (default)\\ g2o,\\ TORO\end{tabular} \\ \hline
\end{tabular}
\end{table}

%\todo{Explain what bold entries in the table mean.}
%done --

\subsection{RTAB-Map}
Real-Time Appearance Based Mapping (RTAB-Map) is a graph-based SLAM approach \cite{labbe_2019} that supports input from RGB-D, Stereo, and lidar sensors. It combines two main algorithms which are \textit{loop closure detector} and \textit{graph optimizer}. The system uses the bag-of-words concept for loop closure detection by determining if the new image comes from a previously visited location or a new location; If the hypothesis of the new image is above a certain threshold, the new location will be added to the map as a new graph constraint. Then, in the background, the map graph is optimized to reduce the drift error in the overall map \cite{Altuntas2017}. To achieve real-time performance for large scale environments, the system has a memory manager that limits and control the number of locations that are used for loop closure detection \cite{labbe_2019}. The system implements two standard odometry methods,  Frame-To-Map (F2M) and Frame-To-Frame (F2F) using 3D visual features. In F2M, the system registers the new frame against upon local map, while the F2F registers the new frame to the last key-frame. RTAB-MAP can generate 2D and 3D occupancy grid map with dense point cloud, which is very useful for robotics applications. Furthermore, there is a full integration of this algorithm in Robot Operating System (ROS) as rtabmap\_ros\footnote{wiki.ros.org/rtabmap\_ros} package.

\subsection{ORB-SLAM2}
ORB-SLAM2 is a feature-based visual SLAM method that can create a sparse 3D map, which can be used with monocular, stereo, and RGB-D cameras to compute the camera trajectory. This algorithm has three main threads running in parallel, which enable real-time performance. The first thread is for tracking the camera pose in the new frames by finding feature matches in the local map. The second thread for local map management and optimization by applying local Bundle Adjustment (BA). The final thread is for loop closure detection and pose-graph optimization; this process is mainly for correcting the accumulated drift \cite{MurArtal2017}. The system uses the bag-of-words DBoW2 concept \cite{Galvez_2012} for place recognition, loop closure, and localization. However, when mapping a large-scale environment, the processing time of loop closure detection and graph optimization will increase as the map grows. This leads to a significant delay when making loop closure corrections after being detected. This system does not generate an occupancy grid map, which makes it difficult to use directly in real robotics applications \cite{labbe_2019}. ORB-SLAM2 does not have full integration with ROS, it only subscribes to the camera topics and there are no output topics. However, it offers a visualizer for the trajectory and a sparse point cloud.

\subsection{LIO-SAM}
LIO-SAM is a factor graph tightly-coupled lidar inertial odometry via smoothing and mapping system \cite{liosam2020shan}. The main input sensors to the system are 3D lidar and 9-axis IMU, but it can also use data from GNSS sensors for absolute measurement and map correction. LIO-SAM estimates lidar motion during the scan by using the raw IMU data. Then, for point cloud de-skewing, it assumes a nonlinear motion model. The novelty of this algorithm is that it uses the idea of keyframes and sliding windows from visual SLAM systems. In this way, the scan registration is performed at the local window scale instead of the global map improving the real-time performance significantly. On the other hand, the old scans are used for pose optimization. The IMU data is critical for this system to work properly. LIO-SAM is fully integrated into ROS. The generated output map and trajectory could be saved on a disk after finishing the scan. 

%there are no published papers about this algorithm!! 
\subsection{StaticMapping}
StaticMapping is a 3D lidar SLAM algorithm with optional IMU, odometry and GNSS inputs \cite{edward_liu_2021}. The back end of this algorithm uses M2DP \cite{He2016} global descriptor for loop closure detection, and iSAM2 \cite{Kaess2011} for smoothing and mapping. This is an offline map-building algorithm from the recorded data. 

%\todo{remove rosbag so it is more general}

% having perfect repeated pattern may not help orb slam
% todo: I need to add the following two experiments: one with slop, and the other one with the different stage of plant growth 

\section{Evaluation} \label{evaluation}
This section describes our simulation environment, the different scenarios to test the algorithms, and the metrics we used for evaluation. Finally, we present and discuss the results. 

\subsection{Environment}
\begin{figure}[!ht]
    \centering
    \begin{subfigure}[b]{0.30\textwidth}
        \includegraphics[width=\textwidth, angle=180]{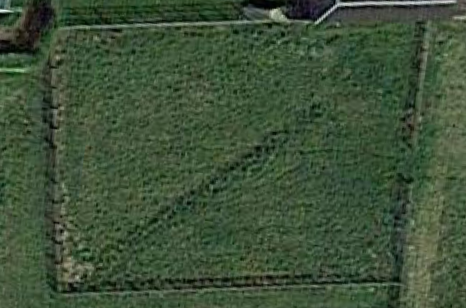}
        \caption{Real field}
        \label{}    
    \end{subfigure}
    \begin{subfigure}[b]{0.28\textwidth}
        \includegraphics[width=\textwidth, angle=180]{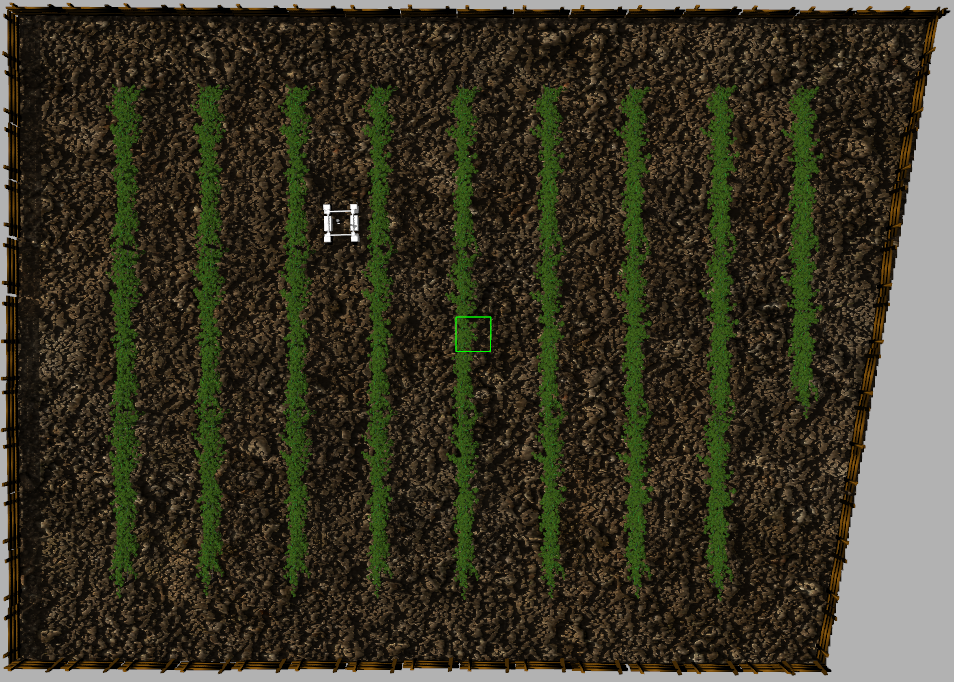}
        \caption{Simulation}
        \label{}    
    \end{subfigure}
    \begin{subfigure}[b]{0.40\textwidth}
        \includegraphics[width=\textwidth]{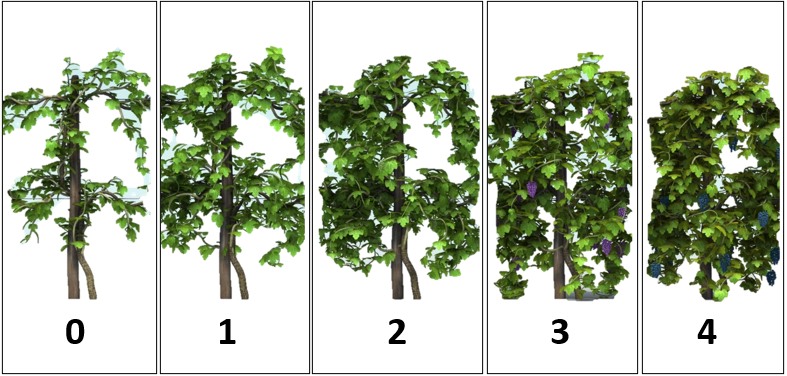}
        \caption{Growth stages}
        \label{}    
    \end{subfigure}
    \begin{subfigure}[b]{0.52\textwidth}
        \includegraphics[width=\textwidth, angle=0]{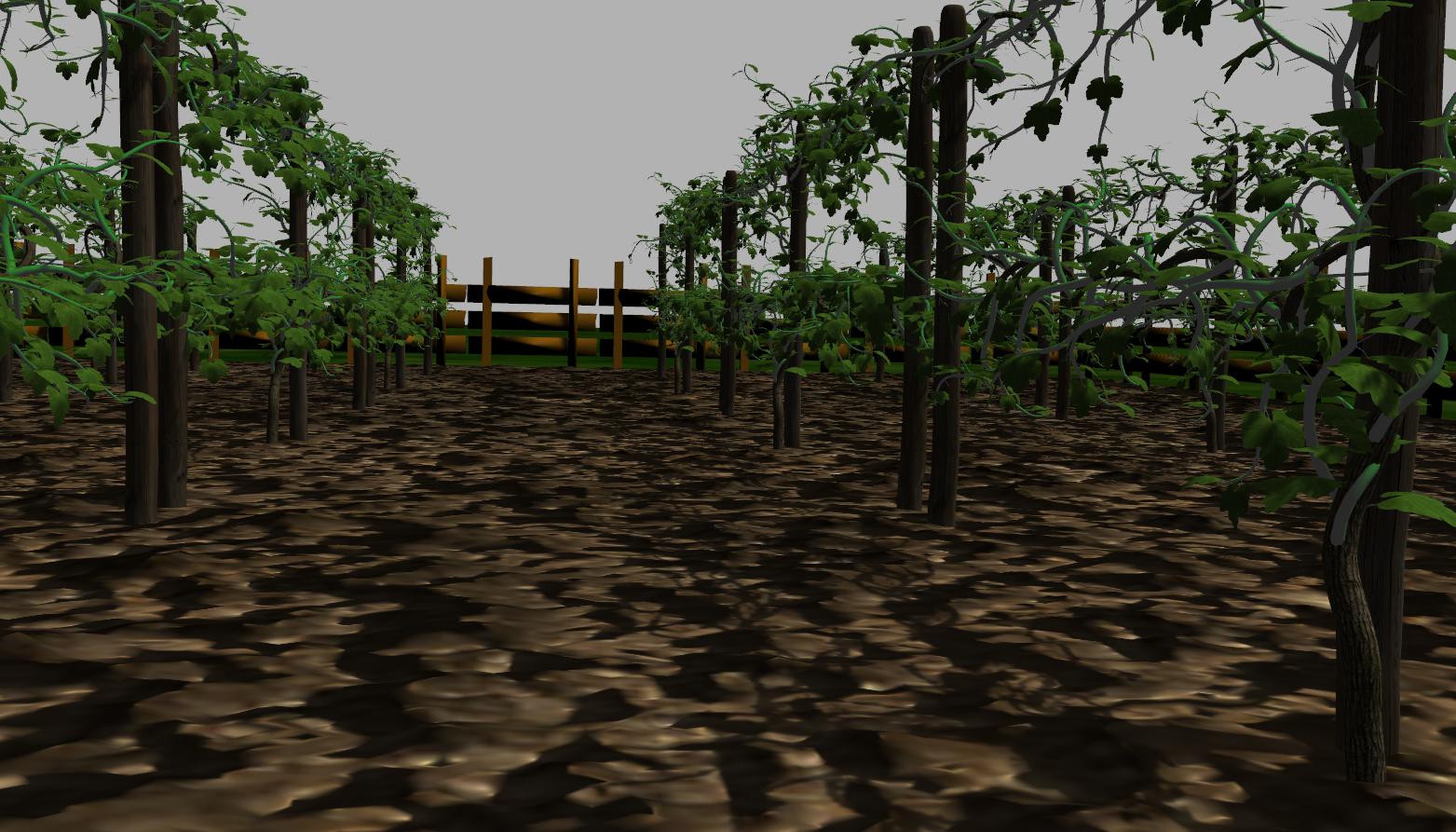}
        \caption{View from robot front camera}
        \label{}    
    \end{subfigure}     
    \begin{subfigure}[b]{0.45\textwidth}
        \includegraphics[width=\textwidth, angle=0]{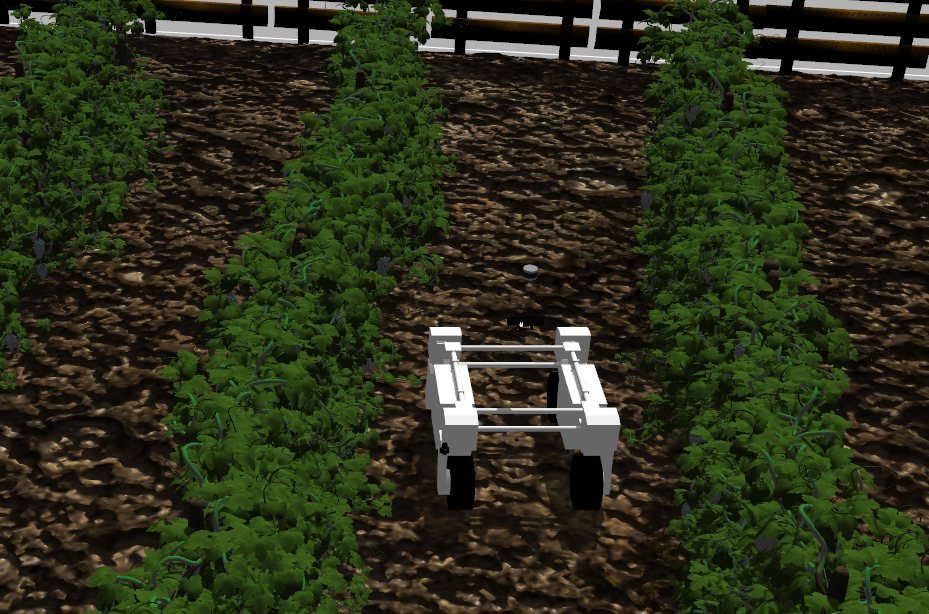}
        \caption{Centric view in final growth stage}
        \label{}    
    \end{subfigure}       
        \caption{Digital twin of the vineyard environment.}
        \label{fig:digital_twin}
\end{figure}

We created a digital twin of an actual vineyard located at the University of Lincoln Riseholme campus under ROS/Gazebo as shown in Figure~\ref{fig:digital_twin}. The virtual environment offers realistic uneven terrain, plus multiple growth stages of the vine plants and the crops. The vineyard has nine rows that are $18$ m long with a $3$ m distance between the rows. The mobile robot used is Thorvald produced by Saga Robotics\footnote{sagarobotics.com/} equipped with the following sensors: two 2D Hokuyo laser scanners, 3D lidar VLP-16 by Velodyne, ROS-IMU plugin and a Kinect V2 camera. A video of the environment can be seen in the link below\footnote{Field: youtu.be/L9ORZNyWdT0. Uneven terrain: youtu.be/L9ORZNyWdT0}.

%not sure about including those figures, I may need to remove them! 

\subsection{Testing scenarios}
We designed four different scenarios in which we evaluated the 4 different SLAM methods we aim to compare.
\begin{itemize}
  \item Scenario 0 (S0): Move in a straight line. That is mainly to evaluate the drift of the generated trajectory. Trajectory length is $25$ m. 
  \item Scenario 1 (S1): Send the robot to inspect a row and get back from the same row. Trajectory length is $48.6$ m.
  \item Scenario 2 (S2): Send the robot to inspect a row and get back from the adjacent row. Trajectory length is $54.6$ m.
  \item Scenario 3 (S3): Inspect multiple rows. That is to simulate a real-life inspection scenario with multiple loop closures. Trajectory length is $101.2$ m.
\end{itemize}

%\todo{Be consistent with units. It is typicall number followed by space followed by SI unit without italics.}

\subsection{SLAM algorithms configurations}
As mentioned earlier, the tested SLAM systems have some integration with ROS, either partially like ORB-SLAM2, or fully as in LIO-SAM. The configuration file for each algorithm has been modified to work with our setup. (a) For ORB-SLAM2, we used the RGBD configuration and modify the camera parameters in the setting file based on the used camera. (b) In RTAB-MAP, the configurations are passed through the launch file. We used the \texttt{rgbd\_sync} node to synchronize the RGB-D camera data before passing them to \texttt{rtabmap\_ros} node. A laser scan data was passed to construct a 2D occupancy grid map. (c) StaticMapping is an offline system, it can only construct the map from recorded data. The default parameters were used and \texttt{accumulate\_cloud\_num} was set to $1$. (d) For LIO-SAM we used the default configurations but disabled the GNSS optimization.

%\todo{To me this last paragraph is a description of the experimental conditions/scenario rather than results. I would move that to earlier part in this section and perhaps reduce its length as it is mainly implementation stuff.}

\subsection{Metrics}
The output trajectory of a SLAM system can be evaluated by finding the absolute distance between the estimated trajectory and the ground truth. The Absolute Trajectory Error (ATE) is defined as the average deviation from the ground truth trajectory\cite{Prokhorov2019}. 
\begin{equation*} \displaystyle \mathrm{ATE}_{rmse}=\left(\frac{1}{n}\sum_{i=1}^{n}\Vert \mathrm{trans} (Q_{x}^{-1}SP_{i})\Vert^{2}\right)^{\frac{1}{2}} \end{equation*},
where $i$ is the time sample or frame, $SP_{i}$ is the spatial translation at time $i$, $trans$ is the translation error, and $Q_{x}$ is the ground truth pose. 
To find the statistical metrics of ATE, we used the open-source library \textit{evo}\footnote{github.com/MichaelGrupp/evo} to calculate the following metrics: Maximum, Mean, Median, Root Mean Square Error (RMSE), and Standard deviation (Std). 

%\todo{Check the equation: I think you mean $trans(Q)SP_i$. Maybe easier to introduce $T_{est}$ and $T_{gt}$. Also is it a dot product or difference?}
%I got this equation from "Measuring robustness of Visual SLAM" paper, originally they had the error term which is E_{i}:=Q_{x}^{-1}SP_{i}, then the ape equation used the E_{i}, I have just copied that into the main equation instead of having 2

\subsection{Results}
\begin{figure}[!ht]
    \centering
    \begin{subfigure}[b]{0.495\textwidth}
        %\centering
        \includegraphics[width=\textwidth]{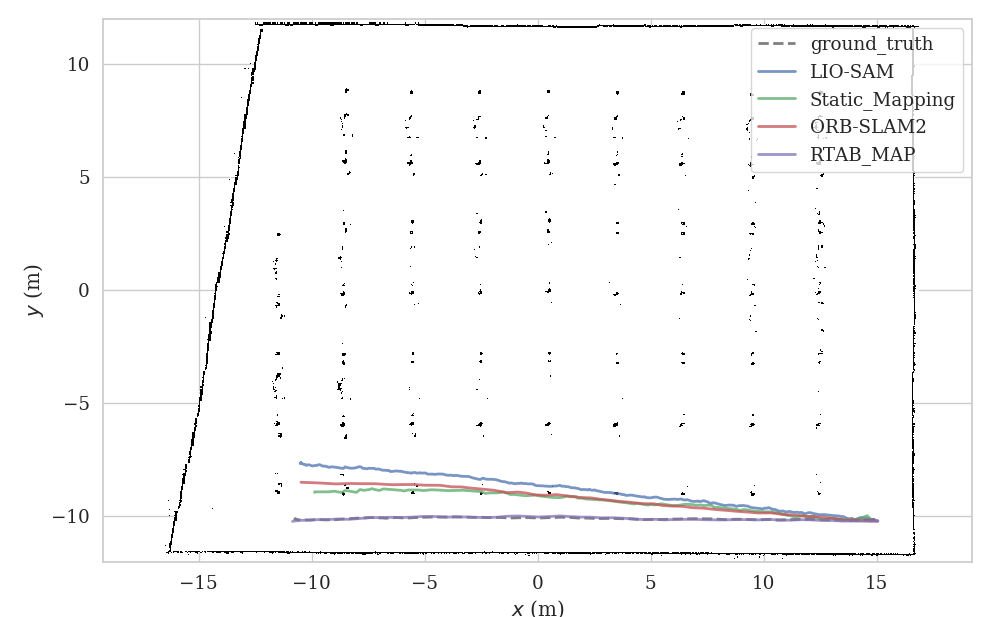}
        \caption{S0}
        \label{fig:scenario 0}    
    \end{subfigure}
    \begin{subfigure}[b]{0.495\textwidth}
        %\centering
        \includegraphics[width=\textwidth]{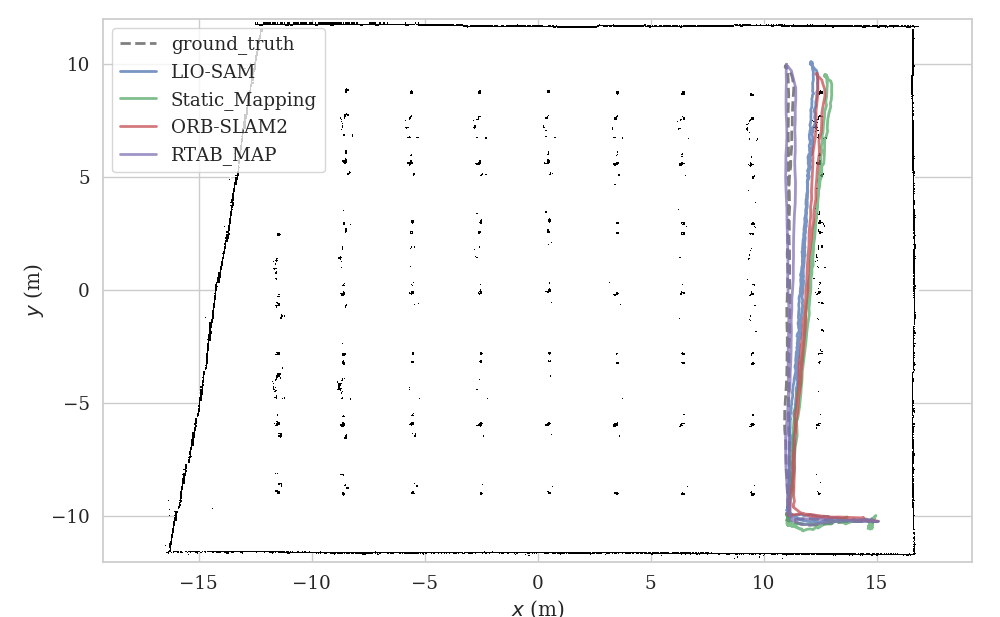}
        \caption{S1}
        \label{fig:scenario 1}    
    \end{subfigure}
    \begin{subfigure}[b]{0.495\textwidth}
        \centering
        \includegraphics[width=\textwidth]{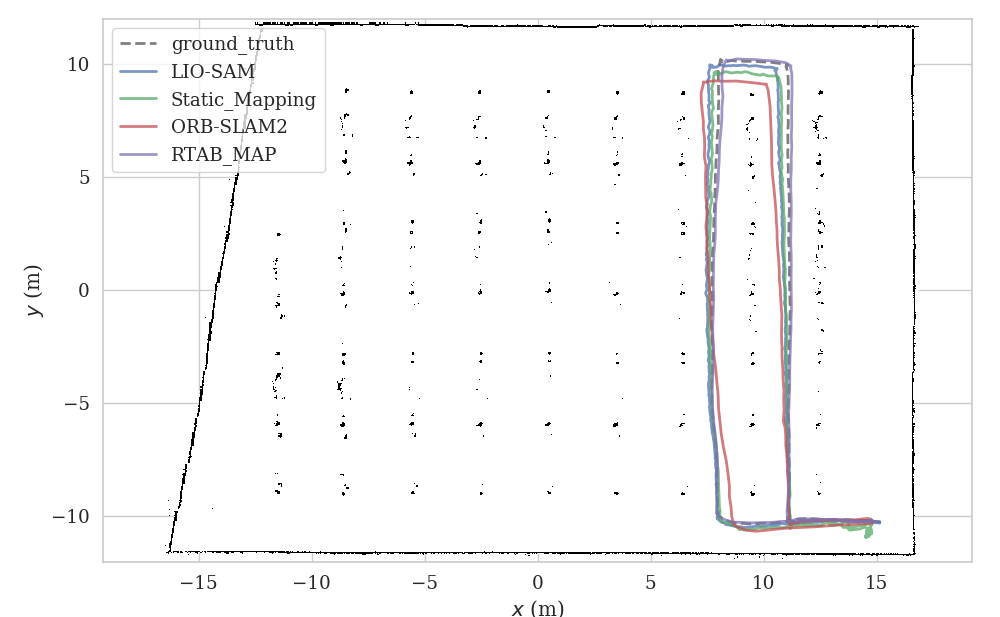}
        \caption{S2}
        \label{fig:scenario 2}    
    \end{subfigure}
    \begin{subfigure}[b]{0.495\textwidth}
        \centering
        \includegraphics[width=\textwidth]{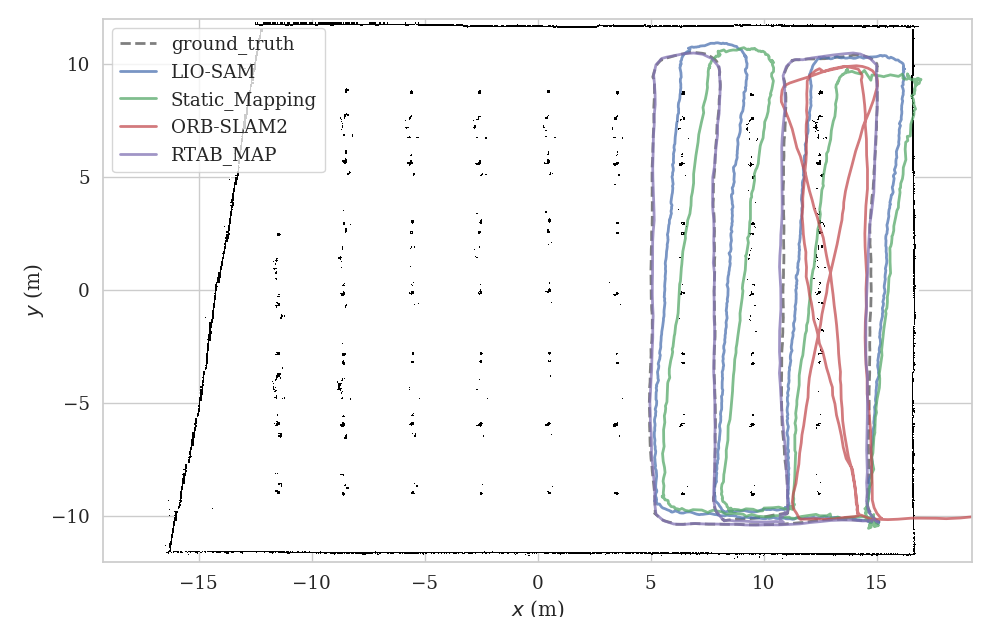}
        \caption{S3}
        \label{fig:scenario 3}    
    \end{subfigure}    

        \caption{Output trajectory from various SLAM systems in different test scenarios.}
        \label{fig:test_scenarios}
\end{figure}

The error metrics of the experiments are summarized in Table \ref{table:results}. Figure \ref{fig:test_scenarios} shows the output trajectories of the four algorithms compared to the ground truth. As it can be observed from Table \ref{table:results} RTAB achieves superior performance with the lowest RMSE across all scenarios compared to other systems.

% \usepackage{multirow}
\begin{table}[]
\caption{ATE statistical metrics for various SLAM systems on 4 different scenarios}
\label{table:results} 
\begin{tabular}{|c|llll|llll|llll|llll|}
\hline
 & \multicolumn{4}{c|}{\textbf{S0}} & \multicolumn{4}{c|}{\textbf{S1}} & \multicolumn{4}{c|}{\textbf{S2}} & \multicolumn{4}{c|}{\textbf{S3}} \\ \cline{2-17} 
\multirow{-2}{*}{} & \multicolumn{1}{c|}{\rotatebox{90}{LIO-SAM }} & \multicolumn{1}{c|}{\rotatebox{90}{STATIC }} & \multicolumn{1}{c|}{\rotatebox{90}{ORBSLAM2 }} & \multicolumn{1}{c|}{\rotatebox{90}{RTAB-MAP}} & \multicolumn{1}{c|}{\rotatebox{90}{LIO-SAM }} & \multicolumn{1}{c|}{\rotatebox{90}{STATIC }} & \multicolumn{1}{c|}{\rotatebox{90}{ORBSLAM2 }} & \multicolumn{1}{c|}{\rotatebox{90}{RTAB-MAP}} & \multicolumn{1}{c|}{\rotatebox{90}{LIO-SAM }} & \multicolumn{1}{c|}{\rotatebox{90}{STATIC }} & \multicolumn{1}{c|}{\rotatebox{90}{ORBSLAM2 }} & \multicolumn{1}{c|}{\rotatebox{90}{RTAB-MAP}} & \multicolumn{1}{c|}{\rotatebox{90}{LIO-SAM }} & \multicolumn{1}{c|}{\rotatebox{90}{STATIC }} & \multicolumn{1}{c|}{\rotatebox{90}{ORBSLAM2 }} & \multicolumn{1}{c|}{\rotatebox{90}{RTAB-MAP}} \\ \hline
MAX    &1.65 & 1.55 & 1.62 & \textbf{0.20} & 1.32 & 1.85 & 1.42 & \textbf{0.08} & 0.83 & 0.79 & 1.34 & \textbf{0.20} & 0.77 & 2.64 & 8.95 & \textbf{0.12} \\ \cline{1-1}
MEAN   & 1.00 & 0.97 & 0.81 & \textbf{0.08} & 0.53 & 0.77 & 0.86 & \textbf{0.05} & 0.20 & 0.42 & 0.69 & \textbf{0.12} & 0.36 & 1.23 & 4.35 & \textbf{0.07} \\ \cline{1-1}
MEDIAN & 1.11 & 1.03 & 0.81 & \textbf{0.05} & 0.40 & 0.29 & 1.04 & \textbf{0.05} & 0.18 & 0.40 & 0.78 & \textbf{0.12} & 0.34 & 0.98 & 3.91 & \textbf{0.06} \\ \cline{1-1}
%MIN    & 0.00 & 0.15 & 0.04 & 0.02 & 0.00 & 0.02 & 0.02 & 0.01 & 0.00 & 0.14 & 0.00 & 0.01 & 0.00 & 0.03 & 0.20 & 0.01 \\ \cline{1-1}
RMSE   & 1.13 & 1.07 & 0.95 & \textbf{0.09} & 0.68 & 1.04 & 0.96 & \textbf{0.05} & 0.22 & 0.45 & 0.84 & \textbf{0.14} & 0.40 & 1.51 & 5.22 & \textbf{0.07} \\ \cline{1-1}
STD    & 0.53 & 0.46 & 0.50 & \textbf{0.05} & 0.43 & 0.70 & 0.42 & \textbf{0.01} & 0.10 & 0.17 & 0.47 & \textbf{0.06} & 0.19 & 0.88 & 2.90 & \textbf{0.03} \\ \hline
\end{tabular}
\end{table}
 
\begin{figure}[!ht]
    \centering
    \begin{subfigure}[b]{0.22\textwidth}
        %\centering
        \includegraphics[width=\textwidth, angle=179]{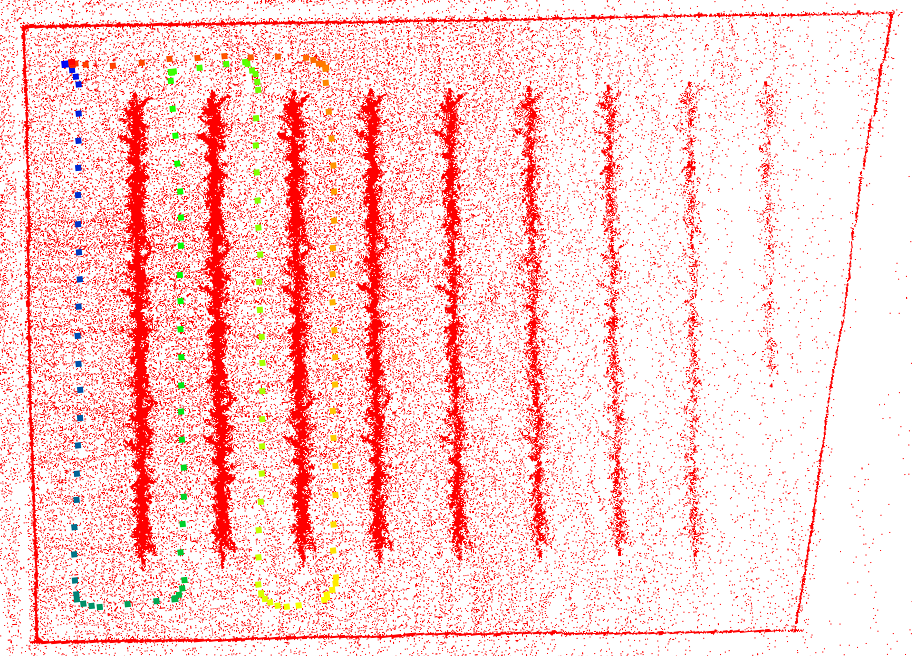}
        \caption{LIO-SAM}
        \label{fig:lio_multi}    
    \end{subfigure}
    \begin{subfigure}[b]{0.22\textwidth}
        %\centering
        \includegraphics[width=\textwidth, angle=180]{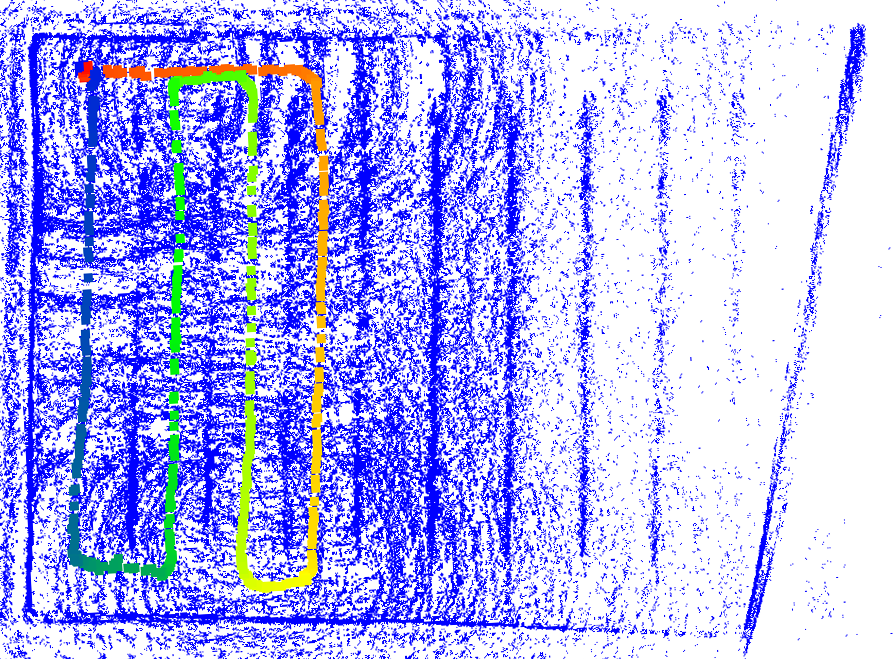}
        \caption{StaticMapping}
        \label{fig:static_multi}    
    \end{subfigure}
    \begin{subfigure}[b]{0.24\textwidth}
        \centering
        \includegraphics[width=\textwidth, angle=180]{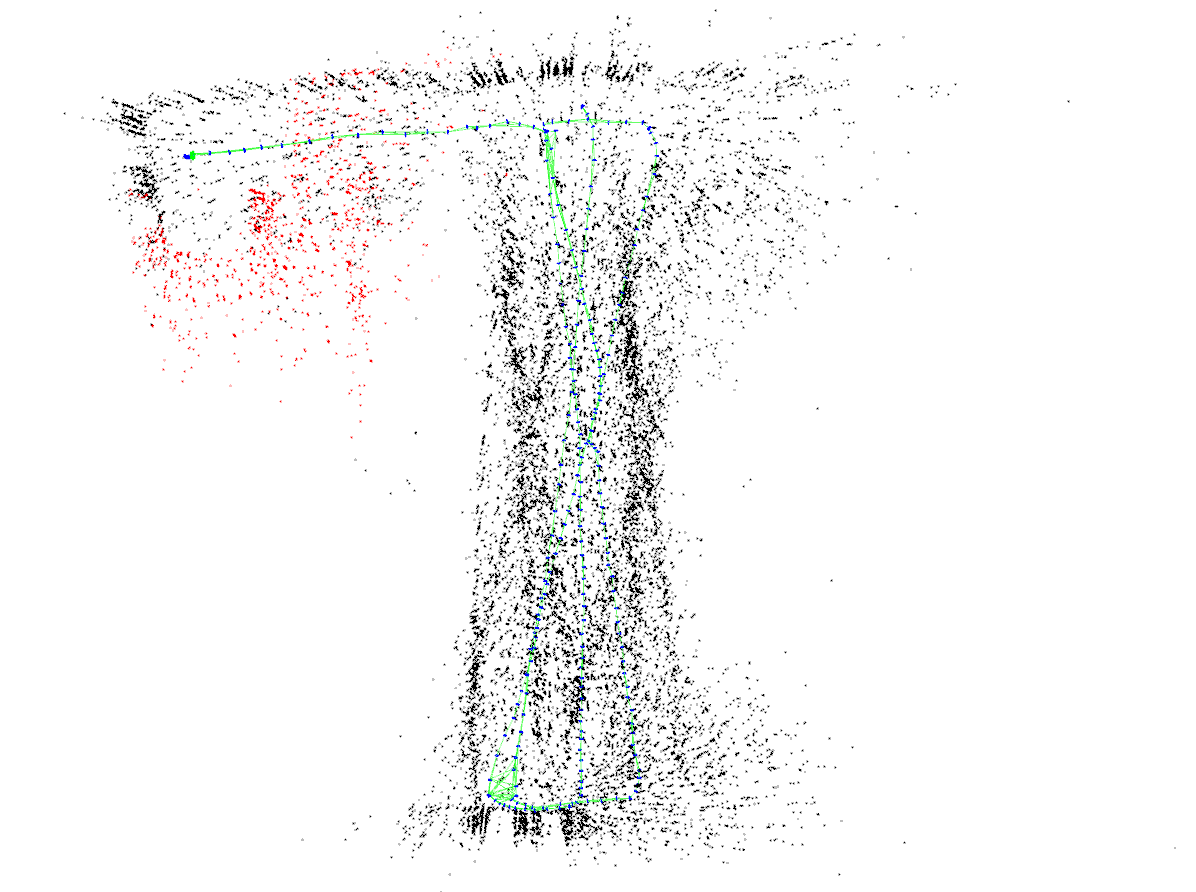}
        \caption{ORB-SLAM2}
        \label{fig:ORB_multi}    
    \end{subfigure}
    \begin{subfigure}[b]{0.24\textwidth}
        \centering
        \includegraphics[width=\textwidth, angle=180]{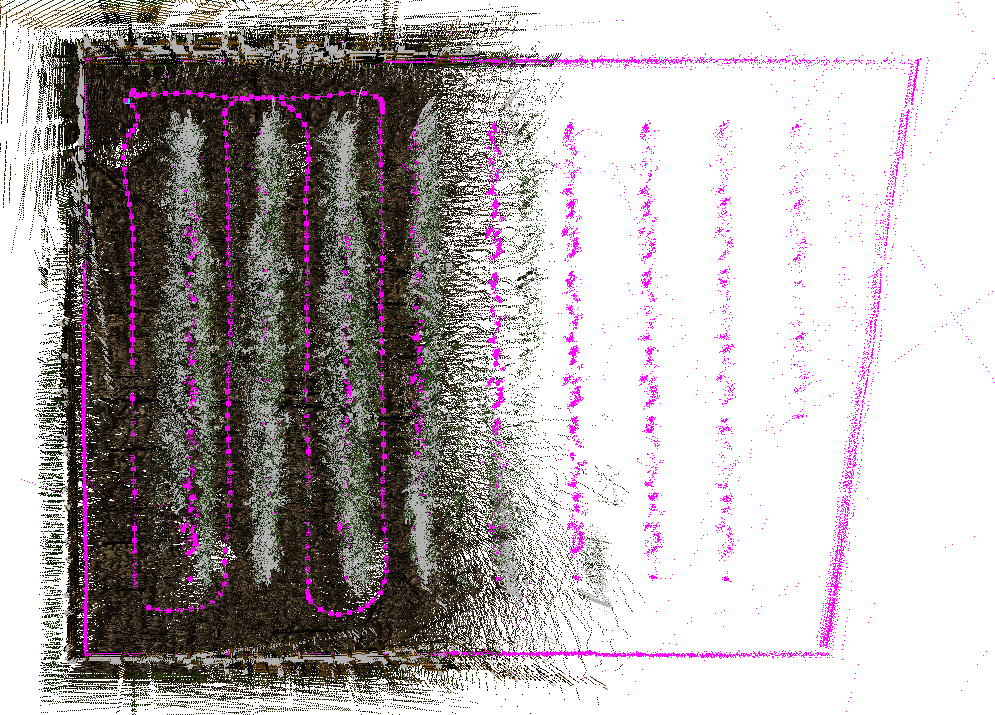}
        \caption{RTAB-MAP}
        \label{fig:rtab_multi}    
    \end{subfigure}    
        \caption{Output maps and estimated trajectory from different SLAM system for scenarios 3}
        \label{fig:output_result_s3}
\end{figure}

In the first scenario S0, all SLAM methods except RTAB-MAP reported big drift in the estimated trajectory, with a maximum value of $1.65$ m from LIO-SAM, while RTAB-MAP has the minimum drift with a maximum value of $0.20$ m and RMSE of $0.09$ m. For scenario S1, RTAB-MAP has the minimum drift and smallest RMSE with $0.05$ m, the performance of the remaining algorithms were close. There is no loop closure in this scenario even though the robot got back to the same place, which is due to the difference in the camera viewpoint. In scenario S2, the second-best algorithm after RTAB-map is LIO-SAM, followed by StaticMapping, and finally ORB-SLAM2 with the largest RMSE of $0.84$ m. The final scenario is the most interesting and realistic one, where the robot traverses multiple rows before going back to its starting point. This scenario with multiple loop closures represents a very interesting benchmark for the algorithms, as represented in Figure~\ref{fig:test_scenarios}. ORB-SLAM2 has failed to create a reliable map, we think that might be due to the repeated visual appearance of the environment; on the other hand, LIO-SAM and StaticMapping did not completely fail. However, the drift of the trajectory is too large to be used for robot navigation. This is because, in a vineyard, the distance between two rows is usually between $2$ to $3$ meters, in which if the mobile robot fails to localize itself accurately it can cause damage to the crops. The map generated for scenario 3 from the different systems is shown in Figure~\ref{fig:output_result_s3}.

%\todo{you can shorten this section by removing long repeatitions of the scenarios - you have introduced them before. you can simply say: In scenario S0, all SLAM methods except RTAB-MAP reported big drift in the estimated trajectory... Also the same remark about the SI units and notation.}

%\todo{Point to the table and result here.}
Even though both ORB-SLAM2 and RTAB-MAP are using the bag-of-words concept for loop closure detection, the pose-graph optimization is different. In ORB-SLAM2, global bundle adjustment is used for the pose optimization process after loop closure. So, since there are lots of visual features shared between the keyframes due to similar appearance, this algorithm fails to estimate reliable trajectory as represented in figure \ref{fig:scenario 3} and table \ref{table:results}. On the other hand, RTAB-MAP is much more robust to false loop closures, since it checks the transformation in the graph after the optimization process, if the translation variance was too large, the loop closure is rejected. The lidar methods did not suffer from repeated feature issue, due to the large field of view of the 3D lidar. 
\section{Conclusion} \label{conclusion}
In this work, we compared four state-of-the-art visual and 3D lidar SLAM algorithms in a challenging simulated vineyard environment with uneven terrain. The main challenge for the visual SLAM system in such an environment is represented by a repeated pattern of appearance and less distinct features. This may result in false loop closures. The state of the art ORB-SLAM2 failed when the robot moved across multiple rows with multiple loop closures, we think this may be due to an identical visual appearance between vineyard rows. RTAB map is much more robust to invalid loop closure. The trajectory generated from the RTAB-MAP algorithm was the most accurate in our test scenarios. The estimated trajectories from LIO-SAM and StaticMapping suffer from big drift but the trajectories shape is acceptable, we believe in real-world tests, this drift could be fixed with the availability of GNSS signal. 
For future work, we will add some unique features within vineyard rows to test the reliability of ORB-SLAM2. Those unique features could be to have some variation in the plant models instead of having them identical across all the rows. In addition, we would like to test the localization ability within the generated maps from the SLAM systems. Finally, we will test those methods in a real vineyard.

%\todo{provide some examples of these unique features} Test the localization ability within the generated maps from the SLAM systems. \todo{make it into a sentence: e.g. In addition, we would like to test the localization ability...} Finally, we will test those methods in a real vineyard.

%\todo{Conclusions should shortly summarise the findings and contributions. You can move some of your analysis to the result section and here only concentrate on the big picture.}
\subsection*{Acknowledgement}
This work has been supported by the European Commission as part of H2020 under grant number 871704 (BACCHUS). 

\bibliographystyle{splncs}
\bibliography{ref}

\end{document}